1# Emotional State Categorization from Speech: Machine vs. Human

Arslan Shaukat, and Ke Chen, *Senior Member, IEEE*

**Abstract**—This paper presents our investigations on emotional state categorization from speech signals with a psychologically inspired computational model against human performance under the same experimental setup. Based on psychological studies, we propose a multistage categorization strategy which allows establishing an automatic categorization model flexibly for a given emotional speech categorization task. We apply the strategy to the Serbian Emotional Speech Corpus (GEES) and the Danish Emotional Speech Corpus (DES), where human performance was reported in previous psychological studies. Our work is the first attempt to apply machine learning to the GEES corpus where the human recognition rates were only available prior to our study. Unlike the previous work on the DES corpus, our work focuses on a comparison to human performance under the same experimental settings. Our studies suggest that psychology-inspired systems yield behaviours that, to a great extent, resemble what humans perceived and their performance is close to that of humans under the same experimental setup. Furthermore, our work also uncovers some differences between machine and humans in terms of emotional state recognition from speech.

**Index Terms**—Acoustic features, activation-evaluation space, emotional speech categorization, machine vs. human performance, multistage categorization model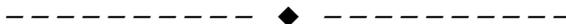

## 1 INTRODUCTION

EMOTION plays a critical role in human communication. In some situations, it is even more important than the logical information contained in speech [1]. Automatic emotional state categorization from speech signals refers to establishing a recognition system to categorize emotional states from speech signals. A number of efforts have been made for such tasks. This includes recording and collection of emotional speech corpora used to analyze the emotional information carried in speech signals [2], [3], [4]. The exploration of acoustic features characterizing different emotional states has also been done by researchers. This includes both utterance-level and segment-level approaches to feature extraction [5], [6], [7], [8]. The applications of state-of-the-art machine learning techniques in automatic categorization have been explored [7], [9]. Feature selection techniques have been applied to discover possible acoustic features responsible for emotional speech [7], [10]. Human performance of recognizing emotions has been compared to the machine performance in [9], [11], [12]. To our best knowledge, however, there are few studies that investigate the performance of an automatic categorization system against humans under the same experimental setup.

In this paper, we first present a multistage emotional state categorization strategy inspired by psychological studies in emotion and its underlying technique. The strategy allows for establishing an automatic categorization system flexibly for a given task. We apply the proposed strategy to the GEES, a Serbian emotional speech corpus [3] and the DES, a Danish emotional speech corpus [4]. For the GEES corpus, there is only human emotion recognition performance reported prior to our study [13]. For the DES corpus, no previous studies in automatic categorization address their performance under the same experimental settings of human listening tests [4]. As a result, our studies investigate the performance and behaviour of an automatic categorization system against those of humans under the same experimental setup for the first time. In our work, we use three acoustic representations in our systems to look into their roles in comparison to the human's performance. Moreover, we also investigate the automatic categorization performance with only the universal acoustic features, irrespective of linguistic and semantic factors, discovered with innovative feature selection strategy in our recent work [14]. In general, our automatic categorization systems exhibit similar behaviors as observed in human listening tests with a variety of acoustic representations.

The paper is organized as follows. Sect. 2 proposes a generic multistage categorization strategy and its underlying technique. Sect. 3 applies the technique to create corresponding automatic categorization systems for the GEES and the DES corpora. Sect. 4 describes our experimental methodology. Sect. 5 reports detailed comparative results on the two corpora: machine vs. human. The last section draws conclusions.

## 2 MODEL DESCRIPTION

This section describes the foundation that leads to the generic multistage categorization strategy and a model derived from such a strategy. The strategy has been in-

————————————————

- *A. Shaukat is with the School of Computer Science, The University of Manchester, Manchester, M13 9PL, U.K. E-mail: Arslan.Shaukat@postgrad.manchester.ac.uk.*
- *K. Chen is with the School of Computer Science, The University of Manchester, Manchester, M13 9PL, U.K. E-mail: Chen@cs.manchester.ac.uk.**Manuscript received (insert date of submission if desired). Please note that all acknowledgments should be placed at the end of the paper, before the bibliography.*xxxx-xxxx/0x/$xx.00 © 200x IEEE

spired from human psychology and hence a review of such inspiration is presented as well.

## 2.1 Psychologically Inspired Strategy

In psychology, universal emotional states can be arguably grouped into higher level dimensions [15]. This has also been supported by activation-evaluation space proposed in [16]. In this space, all universal emotional states are positioned into two dimensions; i.e., activation and evaluation, as illustrated in Fig. 1. Evaluation refers to an emotional state as to how positive or negative the emotion is, whilst activation measures its excitation level in the emotion as to how high or low it is [16].

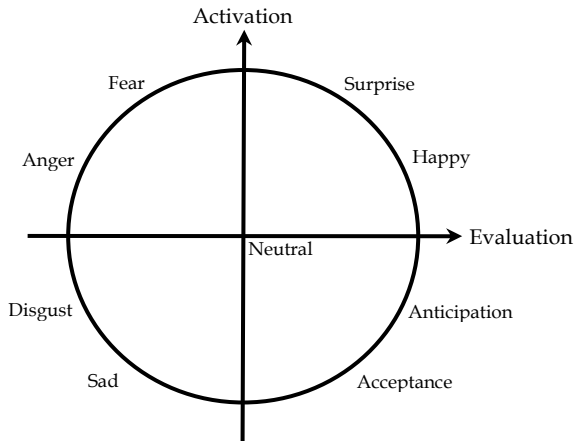

Fig. 1. Activation-evaluation emotion space (adapted from [16]).

Plutchik [17] and Whissell [18] in their independent studies derived the activation and evaluation values for various emotional states. In their study [17], [18], they assigned evaluation values to each emotion ranging from negative extreme 1.1 for guilty to positive extreme 6.6 for delight. Similarly, low activation values around 2.0 are assigned for disinterest and a high activation value of over 6.0 for surprise. In addition, Plutchik [17] also argued that emotional states form a circular arrangement as they are not evenly distributed but form a circular pattern on the activation-evaluation space. He calculated angular measures as to where on the emotion circle, each word lies [16]. As a result, he proposed an emotion wheel shown in Fig. 2. In this wheel, the eight primary emotion dimensions are placed in eight segments arranged in four pairs of opposites.

Two dimensional models reviewed above provide useful ways to describe emotional states and also suggest that humans can effectively identify the emotions at higher categories/taxons. Feeltrace system was developed based on the activation-evaluation emotion space as it allows human users to perceive those emotional contents conveyed in speech [16], [19]. Similarly, the perception of emotions in speech was performed in [20] as how humans categorized emotional states at higher dimensions.

The aim in our work is to categorize discrete emotional states by making use of two dimensional models. To our knowledge, it is unknown how humans categorize the discrete emotion states. Hence, we would like to check if such continuous dimensional models can be applied to discrete emotional states. Both dimensional models [16], [17] suggest that upon listening to an emotional utterance, a subject seems to prefer to determine his/her positive or negative nature as well as its degree of excitation first, and then pay attention to concrete emotional states. In other words, higher dimensions in the activation-evaluation space and the emotion wheel suggest a natural yet easy way to categorize multiple emotional states.

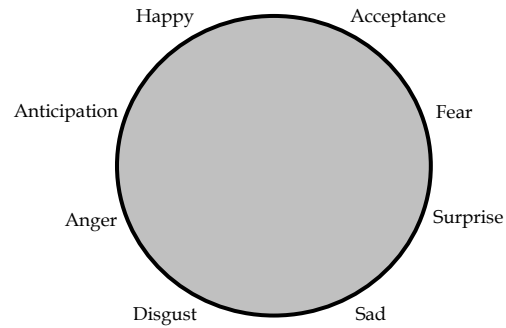

Fig. 2. Plutchik's emotion wheel (adapted from [17]).

Motivated by these previous studies in psychology [15], [17], [18], we firmly believe that an automatic emotional state categorization system should share the same principle; i.e., for automatic categorization, a series of dichotomies always take place in order from high to low taxons defined in the two models. Computationally, dichotomies in a sequential way would be viewed as applying the divide-and-conquer principle to a hard/complex problem by decomposing it into smaller yet simpler sub-problems on different levels for any given emotional state categorization task. This naturally results in a generic multistage categorization strategy to flexibly establish an automatic emotional state categorization system for a given task.

## 2.2 Generalized Categorization Model

In the activation-evaluation emotion space, the higher level taxons along the activation axis correspond to those active and passive states. Similarly, along the evaluation axis, the higher level taxons cover positive and negative states. In our model, we choose the activation axis and first categorize emotions as active and passive states. Since according to listening tests, e.g., two copora used in this paper, human confusions often occur between those states that lie in the same active and passive region of the emotion space, the categorization is based on the activation dimension and the human confusion results.

Emotional states, *happiness*, *sadness*, *anger*, *fear*, *surprise*, and *disgust*, are those so called "Big Six" and have been agreed by most researchers as to be the basic emotions [21]. Such emotional states also appear in the two dimentional models [16], [17]. So our proposed categorization model copes with all six emotional states. In our model, the highest taxon, the dichotomy between "Active" and "Passive" in the activation-evaluation space implemented with a classifier, always first partitions an utterance into two exclusive categories as illustrated in Fig. 3. The "Passive" category is further partitioned by the dichotomy between "Neutral" and "Non-neutral". As a lower taxon, "Non-neutral" is divided by the dichotomy between "Sad" and "Disgust". In the same way, the "Active" category is first partitioned into "Fear" vs. "Non-fear". "Non-fear" is again divided into "Anger" and "Non-Anger" states. "Non-anger" taxon is introduced for the dichotomy between "Happy" and "Surprise" according to the activation-evaluation taxonomy.

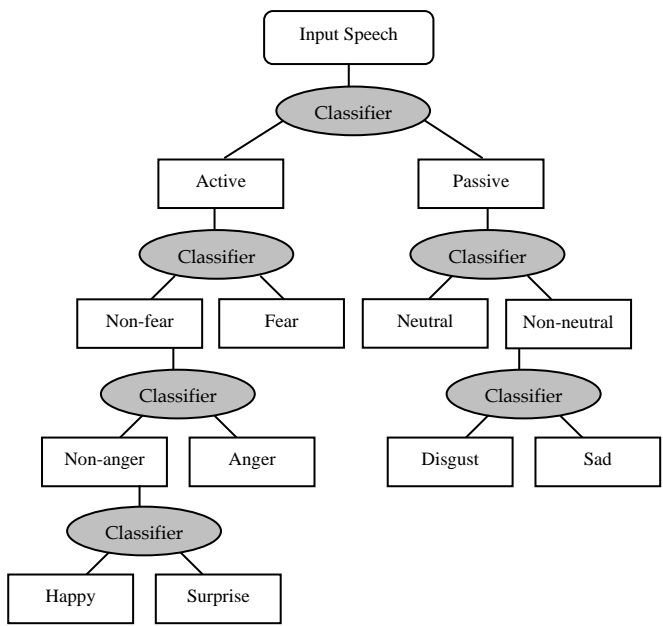

Fig. 3. Generalized multistage categorization model

"Neutral" is neither an "Active" nor a "Passive" state according to the activation-evaluation space [16], but needs to be assigned to a category in our model. According to our analysis on listening tests for the GEES and the DES corpora, we put "Neutral" in the "Passive" state category given the fact that during tests humans mostly confuse "Sad" with "Neutral". Furthermore, previous studies show that such confusion seems quite common based on tests on different corpora as misclassification rates between "Sad"/"Disgust" and "Neutral" states are very high [9], [12], [20]. Thus, in our model, "Neutral" is positioned in the "Passive" category along with "Sad" and "Disgust". We expect that doing so can reduce the misclassification between "Neutral" and those aforementioned states. It is worth stating that it is possible to put "Neutral" in the "Active" category to see how the model behaves whenever it is needed.

Dichotomies for "Fear", "Happy", "Anger" and "Surprise" are formed in the same way by considering the dimensional models and the human listening test results. Four emotional states are in the active dimension of the activation-evaluation and have been often misclassified by human listeners for the two corpora [3], [4]. Therefore, they have been placed under the "Active" category so that the confusion between those emotional states tends to be cleared up. Therefore, our model has a hierarchical structure as shown in Fig. 3. The model is generic in the sense that it can be applied to any emotional speech categorization task irrespective of the number and types of discrete emotional states.

## 3 MULTISTAGE CATEGORIZATION SYSTEMS

In this section, we first describe the two emotional speech corpora and their categorization tasks. Then, a binary classifier, an enabling technique in our system, is presented. Finally, we apply the generalized categorization model to build up system for fulfilling different categorization tasks required by the two corpora.

### 3.1 Emotional Speech Corpora

#### 3.1.1 Serbian Emotional Speech Corpus

The Serbian emotional speech corpus, named GEES, was recorded in an anechoic studio at the Faculty of Dramatic Arts, Belgrade University [3]. The corpus contains 2790 emotional utterances recorded in Serbian. Three actors and three actresses participated in the recording of this corpus. Each speaker uttered 32 words, 30 short sentences, 30 long sentences and a passage composed of 79 words for a single emotional state. The statistics of utterances are phonetically balanced and in agreement with the phonetic statistics of Serbian language [3]. The utterances are labeled by five emotional states, i.e., *happy, sad, anger, neutral* and *fear*. The categorization task is to distinguish these five emotional states from each other. Hence, recognition rates for each emotional state during categorization need to be investigated.

#### 3.1.2 Danish Emotional Speech Corpus

The Danish emotional speech corpus was recorded at Center for Person Komunikation, Aalborg University, Denmark [4]. Four actors, two male and two female, recorded their voices for the corpus. The corpus contains 260 emotional utterances in Danish language. The speakers uttered two single words, nine sentences and two passages for each emotional state, respectively. The script for utterances is semantically neutral; i.e., the text itself expresses no emotional contents [4]. This corpus contains utterances in five emotional states; i.e., *happy, sad, anger, neutral* and *surprise*. Therefore, the categorization task is to classify these five emotional states, which forms a different categorization task from that of the GEES corpus.

### 3.2 Support Vector Machine

As a specific taxon is conceptually defined by the psychological theory [16], [17], [18], we need a powerful underlying dichotomy technique to carry it out for automatic categorization. Theoretical studies in machine learning show that the Support Vector Machine (SVM) [22] along with its kernel-based treatment turns out to be a powerful technique to yield the best generalization performance for binary classification. Therefore, we suggest that the SVM would be an underlying dichotomy technique used in the proposed multistage categorization model to establish a hierarchical dichotomy model for automatic emotional state categorization.

SVM learning maximizes the margin between the support vectors of two classes by finding out a hyperplane of

parameter, $w$ and $b$, as a decision boundary:

$$f(x) = sign(w^T x + b). \quad (1)$$

When data $(x_i, y_i)$ in the feature space are linearly non-separable, a soft margin is introduced by associating a set of positive slack variables $\xi_i$ with the hyperplane. Thus, the optimal hyperplane is obtained by

$$\min_{w,\xi} \frac{1}{2} w^T w + C \sum_{i=1}^{N} \xi_i$$
$$\text{subject to } y_i(w^T \varphi(x_i) + b) \geq 1 - \xi_i \text{ for i=1,...,N.} \quad (2)$$

Here $C$ is a parameter which determines the trade-off between maximizing the margin and minimizing the classification error. $\varphi$ is a nonlinear operator which maps the input data into a higher dimensional feature space. The primitive problem in (2) can be converted into an equivalent problem in the dual space:

$$\max_{\alpha} L(\alpha) = \sum_{i=1}^{N} \alpha_i - \frac{1}{2} \sum_{i,j=1}^{N} \alpha_i \alpha_j y_i y_j \varphi(x_i) \varphi(x_j)$$
$$\text{subject to } \sum_{i=1}^{N} \alpha_i y_i = 0, \quad 0 \leq \alpha_i \leq C, \quad (3)$$

where $\alpha$ is a Lagrange multiplier for combining two objectives in (2). Thus, solving (3) results in the optimal hyperplane [22]:

$$f(x) = sign(w^T x + b)$$
$$= sign(\sum_{i=1}^{N} y_i \alpha_i \langle \varphi(x), \varphi(x_i) \rangle + b)$$
$$= sign(\sum_{i=1}^{N} y_i \alpha_i K(x, x_i) + b)$$
$$= sign(\sum_{i \in SV} w'_i K(x, x_i) + b), \quad (4)$$

where $w'_i = \alpha_i y_i$ is non-zero and $K(x, x_i)$ is the kernel function.

### 3.3 System Description

As two corpora concern different categorization tasks, we apply the generalized categorization model described in Sect. 2 to establish two systems for different tasks.

To meet the requirement of the GEES, the model is derived by pruning two nodes of the generalized categorization model in Fig. 3. The resultant system for the GEES is shown in Fig. 4 where "Surprise" and "Disgust" nodes in the original model, as illustrated in Fig. 3, have been discarded. Similarly, "Fear" and "Disgust" nodes are removed from the original model to create a categorization system for the DES as shown in Fig. 5. As suggested, the SVM technique is used for dichotomy on different levels in these two models. Thus, the application of the multistage categorization strategy leads to two proper multistage systems for automatic categorization on the GEES and the DES corpora.

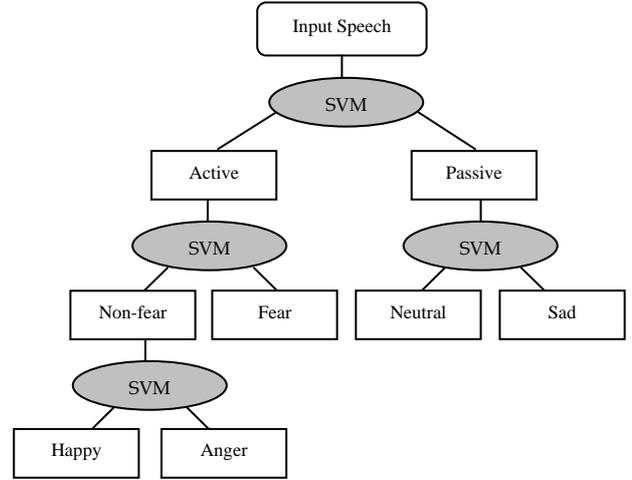

Fig. 4. Structure of multistage categorization for the GEES corpus.

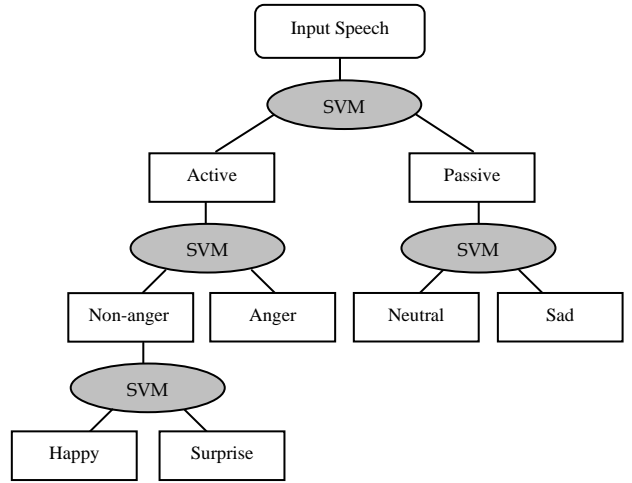

Fig. 5. Structure of multistage categorization for the DES corpus.

It is worth mentioning that there was a similar hierarchical categorization model used for a specific emotional state categorization task. This hierarchical model [23] yielded better performance than traditional methods, which demonstrates the effectiveness of using the divide-and-conquer principle. While our model takes the same principle in the sense of computation, our model design is psychologically inspired, and the SVM classifier used in our systems is theoretically justified to be the right dichotomy technique [22]. Thus, we believe that the generalized model in Fig. 3 form a well qualified system to be employed against human's performance for categorizing emotions under the same experimental setup.

## 4 EXPERIMENTAL METHODOLOGY

In this section, we first present our experimental setting in contrast to the human counterparts and then describe different kinds of acoustic features used in our experiments as they reflect possible ways of automatically re-cognizing emotional states from speech signals.

## 4.1 Experimental Setup

In order to design comparable experiments, we first review the experimental methodology used for human tests and then present our experimental setting.

For the GEES corpus, the detailed categorization performance of 30 normal-hearing listeners are available [3]. In the test, all the listeners were asked to listen to the emotional speech uttered by only one speaker at a time and they were allowed to replay the same utterance several times. Once they finished listening, they made a decision for the given utterance. Note that listeners were not allowed to go back from the middle of an utterance and alter a decision once it had been made for an utterance. Moreover, the test was done on sup-corpora for every individual speaker separately; i.e., listeners were presented with utterances of each speaker separately. The performance was also reported for utterances of various lengths excluding passages. In the listening test, human listeners were not presented with passages to recognize emotions [3].

For the DES corpus, results of 20 normal-hearing listeners are available. Exactly the same experiment setting for the GEES was used in the DES. In the DES, four listening tests have been performed, one for each of the actors in the corpus, which results in four separate results. The human performance is available for individual speaker sub-corpora, three types of utterances and sub-corpora in terms of speakers of different ages [4].

For comparison, we conduct our experiments for the machine performance under the same experimental setting of human listening tests for the GEES corpus. The machine recognition experiments have been performed on the utterances of each speaker separately. As a result, our experiments include individual speaker sub-corpora, different utterance sub-corpora and gender sub-corpora. We have excluded the passages as well from the GEES corpus to make our experimental setup as same as the human listening test.

In [9], [24], [25], automatic emotional speech recognition results are compared to human recognition results for the DES corpus. In our work, however, we exactly follow the experimental setup used in human listening tests. Thus, we have performed four separate automatic recognition experiments for the four actors, whereas in [9], [24], [25], no separate experiments for each actor were done and automatic recognition results were reported only on the whole corpus. In the DES corpus, each actor has recorded 65 utterances containing words, sentences and passages. In addition to this, utterances containing target voices are also recorded. In the listening test and our automatic recognition experiments, only 65 utterances for each actor are considered for recognition whereas in [9], [24], [25], they consider the target utterances as part of the corpus and perform their experiments on such kind of corpus which is not according to the settings of listening test. Also in [9], [24], [25], no machine performance for words, sentences and passages was reported. Hence, the automatic categorization performed in these studies did not take into account the human listening test's setting. In summary, we conduct experiments for the DES corpus with individual speaker sub-corpora, different utterance sub-corpora and sub-corpora of speakers at different ages.

In our work, the radial basis function kernel is employed in the SVM classifier for dichotomy at different levels (c.f. Fig. 4 and Fig. 5). For the robustness, we use the *leave-one-out* cross-validation method to evaluate the performance of our automatic categorization systems in all the experiments described below.

## 4.2 Acoustic Representations

Acoustic features especially for emotional state recognition from speech have been studied [5], [6], [7], [23], [25], [26], [27], [28] and a set of acoustic features are identified to be potentially useful for characterizing emotional speech. We employ three types of acoustic representations based on all identified features since there are different types of emotional utterances like words, sentences and passages in the two corpora. We anticipate that different representations may capture useful information for different types of utterances. It is worth clarifying that no matter which representation is used, one decision is merely made for a complete utterance, which resembles the human listening test [3], [4].

Here, we do not claim that acoustic features have any psychological meaning. Humans certainly use different types of acoustic features apart from linguistic and physiological features to recognize vocal emotions. Such statistics of acoustic measurements are chosen simply because they correlate with known evidence in human perception, such as loudness, pitch dynamics etc. So such features are unlikely to have any psychological meaning. Below we briefly describe three different types of acoustic representations used in our experiments.

### 4.2.1 Utterance-Based Representation

An utterance-based representation treats an utterance as a whole, and hence a feature vector or its representation is simply formed for the entire utterance based on feature measures. This representation may be regarded as the global features captured by human listeners. Utterance-based representations used in our work include all 318 possible features, grouped into 11 feature types in terms of feature measures, as summarized in Table 1.

### 4.2.2 Segment-Based Representation

A segment-based approach would block an utterance into several segments and a feature vector is extracted for each segment. This representation tends to capture the critical local features that might be used by humans to identify an emotional utterance. The segment-based technique used in our work was proposed in [8]. Thus, the segment duration and other 295 features extracted based on different measures in Table 1 are used to generate a feature vector for each segment. The collection of feature vectors for segments of an utterance would be used as a representation.

TABLE 1
FULL SET OF VARIOUS ACOUSTIC FEATURES FOR EMOTIONAL SPEECH. UTTERANCE-BASED UNIVERSAL FEATURES ARE HIGHLIGHTED IN BOLD, SEGMENT-BASED UNIVERSAL FEATURES ARE HIGHLIGHTED IN ITALIC AND FEATURES BELONGING TO BOTH SUBSETS ARE HIGHLIGHTED IN BOLD-ITALIC.

| Feature Measure | Full Feature Set |
|---|---|
| Loudness (20 features) | **mean, 25 percentile, 50 percentile, 75 percentile, 25 percentile RMS, 50 percentile RMS, 75 percentile RMS, mean specific loudness band 1 (msl b1)**, msl b2, *msl b3*, **msl b4, msl b5, msl b6, msl b7, msl b8, msl b9, msl b10, msl b11, msl b12, msl b13** [5], [6]. |
| Voice source (28 features) | *25 percentile of $E_e$*, **median of $E_e$**, *75 percentile of $E_e$*, **IQR of normalized $\Delta E_e$**, 25 percentile of $\gamma$, median of $\gamma$, *75 percentile of $\gamma$*, IQR of normalized $\Delta\gamma$, *25 percentile of $\alpha$*, median of $\alpha$, *75 percentile of $\alpha$*, IQR of normalized $\Delta\alpha$, *25 percentile of $\beta$*, median of $\beta$, *75 percentile of $\beta$*, IQR of normalized $\Delta\beta$, 25 percentile of OQ, median of OQ, 75 percentile of OQ, **IQR of normalized $\Delta OQ$**, 25 percentile of $\varepsilon_o$, *median of $\varepsilon_o$*, *75 percentile of $\varepsilon_o$*, IQR of normalized $\Delta\varepsilon_o$, 25 percentile of $\varepsilon_c$, *median of $\varepsilon_c$*, **75 percentile of $\varepsilon_c$**, IQR of normalized $\Delta\varepsilon_c$ [5], [6]. |
| Other voice source (14 features) | **jitter$_{PF}$, max jitter$_{PQ}$, min jitter$_{PQ}$**, shimmer$_{PF}$, ***max shimmer$_{PQ}$***, min shimmer$_{PQ}$, 25 percentile of GNE, *median of GNE*, 75 percentile of GNE, IQR of normalized $\Delta$GNE, **25 percentile of PSP**, *median of PSP*, 75 percentile of PSP, *IQR of normalized $\Delta PSP$* [5], [6]. |
| Harmonicity (14 features) | *median of intrinsic diss. $D_I$*, **range of intrinsic diss. $D_I$**, *median of avg. diss., median of avg. diss. derivative, median of cons. values at interval $\alpha_1^c$, median of highest cons. interval $\alpha_1^c$, median of cons. values at interval $\alpha_2^c$*, median of second highest cons. interval $\alpha_2^c$, *median of avg. cons. peak values*, **median of diss. values at interval $\alpha_1^d$**, median of highest diss. interval $\alpha_1^d$, **median of diss. values at interval $\alpha_2^d$**, median of second highest diss. interval $\alpha_2^d$, **median of avg. diss. peak values** [5], [6]. |
| Fundamental frequency or pitch (44 features) | minima series: **mean**, *max, min, range, var, med*, **1st quartile**, *3rd quartile, iqr, mean abs. val. of derivative.*<br>maxima series: mean, max, *min, range, var, med*, **1st quartile**, 3rd quartile, *iqr*, mean abs. val. of derivative.<br>durations between local extrema series: mean, max, ***min***, range, var, *med*, **1st quartile**, 3rd quartile, iqr, mean abs. val. of derivative.<br>series itself: **mean**, max, ***min***, range, *var, med*, **1st quartile**, 3rd quartile, iqr, mean abs. val. of derivative, *skewness, fraction of voiced F0 above mean*, **range above mean**, *range below mean* [5], [6], [7]. |
| Intensity or energy (40 features) | minima series: mean, max, **min**, range, *var*, *med*, **1st quartile**, 3rd quartile, **iqr**, mean abs. val. of derivative.<br>maxima series: mean, max, min, range, var, med, 1st quartile, 3rd quartile, iqr, **mean abs. val. of derivative.**<br>durations between local extrema series: mean, max, min, range, *var*, med, 1st quartile, 3rd quartile, iqr, mean abs. val. of derivative.<br>series itself: mean, **min**, range, **var**, med, 1st quartile, 3rd quartile, iqr, mean abs. val. of derivative [7]. |
| Low-pass intensity (40 features) | minima series: ***mean, max***, min, *range, var, med*, **1st quartile, 3rd quartile, iqr, mean abs. val. of derivative.**<br>maxima series: ***mean, max***, min, *range, var, med, 1st quartile*, **3rd quartile, iqr, mean abs. val. of derivative.**<br>durations between local extrema series: *mean*, **max, min**, range, var, *med*, 1st quartile, **3rd quartile, iqr, mean abs. val. of derivative.**<br>series itself: ***mean, max***, min, *range, var, med, 1st quartile, 3rd quartile, iqr, mean abs. val. of derivative* [7]. |
| High-pass intensity (40 features) | minima series: mean, max, **min**, range, *var, med*, **1st quartile**, 3rd quartile, iqr, mean abs. val. of derivative.<br>maxima series: mean, **max, min, range**, var, **med**, 1st quartile, 3rd quartile, iqr, **mean abs. val. of derivative.**<br>durations between local extrema series: mean, max, min, range, var, med, 1st quartile, 3rd quartile, iqr, mean abs. val. of derivative.<br>series itself: **mean, max**, ***min***, range, var, med, 1st quartile, 3rd quartile, iqr, mean abs. val. of derivative [7]. |
| Mel-frequency cepstral coefficients (40 features) | minima series: **mean**, *max*, min, *range*, var, ***med, 1st quartile***, 3rd quartile, iqr, ***mean abs. val. of derivative.***<br>maxima series: mean, max, **min**, *range*, var, med, ***1st quartile***, 3rd quartile, **iqr, mean abs. val. of derivative.**<br>durations between local extrema series: mean, max, min, range, **var**, med, **1st quartile**, *3rd quartile*, iqr, mean abs. val. of derivative.<br>series itself: **mean**, *max*, min, *range*, var, ***med, 1st quartile***, **3rd quartile, iqr, mean abs. val. of derivative** [7]. |
| Formant (15 features) | ***mean F1, mean F2***, **mean F3**, *std F1*, **std F2, std F3**, *max F1*, max F2, max F3, *min F1*, min F2, min F3, *range F1*, range F2, range F3 [25], [26]. |
| Duration (23 features) | **mean dur. of aud. segs.**, max dur. of aud. segs., **min dur. of aud. segs., std. of dur. of aud. segs.**, mean dur. of inaud. segs., max dur. of inaud. segs., min dur. of inaud. segs., std. of dur. of inaud. segs., no. of aud. segs., no. of inaud. segs., no. of aud. frames., no. of inaud. frames, longest aud. seg., longest inaud. seg.,<br>ratios of: no. of aud. to inaud. frames, no. of aud. to inaud. segs., **no. of aud. to total no. of frames**, no. of aud. to total no. of segs., no. of aud. frames to no. of aud. segs., total duration of aud. segs. to total duration of inaud. segs., **duration of aud. segs. to total duration of utterance**, duration of inaud. segs. to total duration of utterance, avg. duration of aud. segs. to avg. duration of inaud. segs. [27], [28]. |

### 4.2.3 Combination-Based Representation

While the utterance and the segment based representations are likely to characterize the global and the local features, the combination-based representation tends to combine them for the exploitation of all kinds of features. As there are different combination methods, e.g., [29], we use the simplest method proposed in [8] in our work to investigate the baseline performance by combining global and local features, which is expected to simulate a way that human exploits features for the same task.

### 4.3 Universal Feature Exploration

Humans generally have the ability to recognize emotions from speech even if they do not understand the language [15]. For example, as one in the bad mood is speaking angrily, other people can still judge the emotional state of that person regardless what he utters. It suggests that humans could use some kind of universal acoustic features, irrelevant to linguistics or semantics, to facilitate recognizing emotions. In our recent work [14], we developed a generic feature selection algorithm to explore emotional acoustic features irrespective of linguistics and semantics. The motivation is to find a subset of acoustic features that tends to be universal or applicable to various emotional speech corpora. In this study [14], feature selection techniques are applied to a corpus of a different language, the Berlin emotional speech corpus [2]. The selected features are then used on the GEES and the DES to achieve the recognition rates in the same way as the full acoustic feature set is used.

Universal feature subsets discovered in our work [14] are tabulated in Table 1. The listening test in both the GEES and the DES corpora was done by native listeners. One open question would be how a foreigner who does not know the language and has a different cultural background categorizes emotional states. Although there has been no such human performance on these two corpora so far, our experiments with universal feature subsets would provide such results produced by an automatic categorization system, which provides the baseline performance for future studies.

TABLE 2
EMOTION RECOGNITION RATE FOR INDIVIDUAL SPEAKERS FOR THE GEES CORPUS

| Speakers | Emotions | Machine recognition rate in % | | | | | Machine recognition rate in % | | | | | Machine recognition rate in % | | | | |
|---|---|---|---|---|---|---|---|---|---|---|---|---|---|---|---|---|
| | | Utterance-based representations | | | | | Segment-based representations | | | | | Combination-based representations | | | | |
| | | H | S | A | N | F | H | S | A | N | F | H | S | A | N | F |
| SK | H | **83.70** | 0.00 | 13.04 | 1.09 | 2.17 | **80.43** | 0.00 | 16.30 | 1.09 | 2.17 | **78.26** | 0.00 | 18.48 | 1.09 | 2.17 |
| | S | 0.00 | **90.00** | 0.00 | 5.56 | 4.44 | 0.00 | **97.78** | 0.00 | 2.22 | 0.00 | 0.00 | **94.44** | 0.00 | 4.44 | 1.11 |
| | A | 13.04 | 0.00 | **84.78** | 0.00 | 2.17 | 15.22 | 0.00 | **84.78** | 0.00 | 0.00 | 18.48 | 0.00 | **79.35** | 0.00 | 2.17 |
| | N | 1.09 | 4.35 | 0.00 | **89.13** | 5.43 | 0.00 | 7.61 | 0.00 | **89.13** | 3.26 | 0.00 | 7.61 | 0.00 | **90.22** | 2.17 |
| | F | 3.26 | 1.09 | 0.00 | 3.26 | **92.39** | 9.78 | 2.17 | 1.09 | 8.70 | **78.26** | 2.17 | 0.00 | 1.09 | 1.09 | **95.65** |
| MV | H | **95.65** | 0.00 | 4.35 | 0.00 | 0.00 | **96.74** | 0.00 | 3.26 | 0.00 | 0.00 | **95.65** | 0.00 | 4.35 | 0.00 | 0.00 |
| | S | 0.00 | **95.65** | 0.00 | 1.09 | 3.26 | 0.00 | **89.66** | 0.00 | 0.00 | 10.34 | 0.00 | **96.55** | 0.00 | 0.00 | 3.45 |
| | A | 6.52 | 0.00 | **93.48** | 0.00 | 0.00 | 9.78 | 0.00 | **89.13** | 0.00 | 1.09 | 6.52 | 0.00 | **93.48** | 0.00 | 0.00 |
| | N | 0.00 | 0.00 | 1.09 | **97.83** | 1.09 | 0.00 | 2.17 | 0.00 | **96.74** | 1.09 | 0.00 | 1.09 | 0.00 | **98.91** | 0.00 |
| | F | 0.00 | 4.35 | 0.00 | 1.09 | **94.57** | 0.00 | 5.49 | 0.00 | 0.00 | **94.51** | 0.00 | 3.30 | 0.00 | 0.00 | **96.70** |
| MM | H | **77.17** | 0.00 | 16.30 | 0.00 | 6.52 | **84.78** | 0.00 | 8.70 | 3.26 | 3.26 | **83.70** | 0.00 | 13.04 | 1.09 | 2.17 |
| | S | 0.00 | **90.00** | 0.00 | 5.56 | 4.44 | 0.00 | **92.05** | 0.00 | 5.68 | 2.27 | 0.00 | **89.77** | 0.00 | 3.41 | 6.82 |
| | A | 11.96 | 1.09 | **84.78** | 0.00 | 2.17 | 11.11 | 0.00 | **85.56** | 2.22 | 1.11 | 12.22 | 0.00 | **83.33** | 1.11 | 3.33 |
| | N | 1.09 | 3.26 | 0.00 | **95.65** | 0.00 | 0.00 | 4.35 | 0.00 | **95.65** | 0.00 | 1.09 | 3.26 | 0.00 | **94.57** | 1.09 |
| | F | 10.87 | 4.35 | 5.43 | 1.09 | **78.26** | 22.83 | 2.17 | 11.96 | 4.35 | **58.70** | 9.78 | 2.17 | 4.35 | 3.26 | **80.43** |
| SZ | H | **83.70** | 0.00 | 9.78 | 0.00 | 6.52 | **88.04** | 0.00 | 6.52 | 0.00 | 5.43 | **86.96** | 0.00 | 6.52 | 0.00 | 6.52 |
| | S | 0.00 | **98.91** | 0.00 | 1.09 | 0.00 | 0.00 | **98.91** | 0.00 | 1.09 | 0.00 | 0.00 | **95.65** | 0.00 | 2.17 | 2.17 |
| | A | 8.70 | 1.09 | **90.22** | 0.00 | 0.00 | 11.96 | 0.00 | **80.43** | 2.17 | 5.43 | 11.96 | 0.00 | **81.52** | 2.17 | 4.35 |
| | N | 1.09 | 1.09 | 3.26 | **94.57** | 0.00 | 1.09 | 3.26 | 0.00 | **93.48** | 2.17 | 0.00 | 1.09 | 2.17 | **93.48** | 3.26 |
| | F | 9.78 | 0.00 | 3.26 | 3.26 | **83.70** | 15.22 | 2.17 | 10.87 | 5.43 | **66.30** | 4.35 | 2.17 | 5.43 | 1.09 | **86.96** |
| OK | H | **91.30** | 0.00 | 0.00 | 0.00 | 8.70 | **85.87** | 0.00 | 7.61 | 0.00 | 6.52 | **84.78** | 0.00 | 11.96 | 0.00 | 3.26 |
| | S | 0.00 | **97.83** | 1.09 | 0.00 | 1.09 | 0.00 | **95.65** | 0.00 | 4.35 | 0.00 | 0.00 | **93.48** | 0.00 | 6.52 | 0.00 |
| | A | 0.00 | 0.00 | **96.74** | 0.00 | 3.26 | 15.22 | 0.00 | **82.61** | 1.09 | 1.09 | 15.22 | 0.00 | **83.70** | 0.00 | 1.09 |
| | N | 0.00 | 0.00 | 1.09 | **97.83** | 1.09 | 0.00 | 17.39 | 0.00 | **82.61** | 0.00 | 0.00 | 8.70 | 0.00 | **91.30** | 0.00 |
| | F | 0.00 | 0.00 | 5.43 | 27.17 | **67.39** | 1.09 | 1.09 | 0.00 | 0.00 | **97.83** | 3.26 | 0.00 | 0.00 | 0.00 | **96.74** |
| BM | H | **94.57** | 0.00 | 3.26 | 2.17 | 0.00 | **92.39** | 0.00 | 5.43 | 0.00 | 2.17 | **95.65** | 0.00 | 4.35 | 0.00 | 0.00 |
| | S | 1.09 | **93.48** | 0.00 | 4.35 | 1.09 | 0.00 | **100.00** | 0.00 | 0.00 | 0.00 | 1.09 | **95.65** | 0.00 | 3.26 | 0.00 |
| | A | 3.26 | 0.00 | **94.57** | 2.17 | 0.00 | 4.35 | 0.00 | **89.13** | 4.35 | 2.17 | 3.26 | 0.00 | **92.39** | 4.35 | 0.00 |
| | N | 0.00 | 2.17 | 2.17 | **95.65** | 0.00 | 1.09 | 3.26 | 1.09 | **93.48** | 1.09 | 1.09 | 4.35 | 0.00 | **93.48** | 1.09 |
| | F | 1.09 | 2.17 | 0.00 | 1.09 | **95.65** | 0.00 | 5.43 | 0.00 | 0.00 | **94.57** | 0.00 | 4.35 | 0.00 | 0.00 | **95.65** |

TABLE 3
EMOTION RECOGNITION RATE FOR ALL SPEAKERS FOR THE GEES CORPUS

| Speakers | Emotions | Machine recognition rate in % | | | | | Machine recognition rate in % | | | | | Machine recognition rate in % | | | | |
|---|---|---|---|---|---|---|---|---|---|---|---|---|---|---|---|---|
| | | Utterance-based representations | | | | | Segment-based representations | | | | | Combination-based representations | | | | |
| | | H | S | A | N | F | H | S | A | N | F | H | S | A | N | F |
| All | H | **87.68** | 0.00 | 7.79 | 0.54 | 3.99 | **88.04** | 0.00 | 7.97 | 0.72 | 3.26 | **87.50** | 0.00 | 9.78 | 0.36 | 2.36 |
| | S | 0.18 | **94.31** | 0.18 | 2.94 | 2.39 | 0.00 | **95.67** | 0.00 | 2.22 | 2.10 | 0.18 | **94.26** | 0.00 | 3.30 | 2.26 |
| | A | 7.25 | 0.36 | **90.76** | 0.36 | 1.27 | 11.27 | 0.00 | **85.27** | 1.82 | 1.63 | 11.28 | 0.00 | **85.63** | 1.27 | 1.82 |
| | N | 0.54 | 1.81 | 1.27 | **95.11** | 1.27 | 0.36 | 6.34 | 0.18 | **91.85** | 1.27 | 0.36 | 4.35 | 0.36 | **93.66** | 1.27 |
| | F | 4.17 | 1.99 | 2.36 | 6.16 | **85.33** | 8.15 | 3.09 | 3.99 | 3.08 | **81.69** | 3.26 | 2.00 | 1.81 | 0.91 | **92.02** |

## 5 EXPERIMENTAL RESULTS

In this section, we report our experimental results and carry out an analysis against the human performance on the GEES and the DES based on the full feature set and the universal feature subset.

### 5.1 Results on the Full Feature Set

Now we report results on the two corpora by using three types of acoustic representations formed with the full feature set listed in Table 1 to reflect the machine performance.

#### 5.1.1 GEES Corpus

For the GEES corpus, human listening experiments suggest that "anger" and "happy" states are mostly confused with each other, and so are "sad" and "neutral" states. To a lesser extent, "fear" and "sad" states are also misclassified with each other.

We would report our results in the form of confusion-matrices so that misclassification can be identified and compared with humans'. Tables 2 and 3 present confusion matrices of five emotional states based on the performance of our system for individual sub-corpora and the whole corpus in exactly the same way as reported in [3]. Tables 4 and 5 show the contrastive results by subtracting confusion matrices achieved by our system from the corresponding ones achieved by human listeners in [3], which allows one to see the performance difference explicitly. The notation in Tables 4 and 5 stipulates that a negative/positive number in the diagonal expresses how much the performance of our system is inferior/superior to that of humans, while a negative/positive number in the off-diagonal indicates how much ours is superior/inferior to that of humans.

TABLE 4
CONTRASTIVE RESULTS FOR INDIVIDUAL SPEAKERS FOR THE GEES CORPUS

| Speakers | Emotions | Machine recognition rate in % | | | | | Machine recognition rate in % | | | | | Machine recognition rate in % | | | | |
|---|---|---|---|---|---|---|---|---|---|---|---|---|---|---|---|---|
| | | Utterance-based representations | | | | | Segment-based representations | | | | | Combination-based representations | | | | |
| | | H | S | A | N | F | H | S | A | N | F | H | S | A | N | F |
| SK | H | **-8.68** | -0.27 | 8.94 | -0.48 | 0.78 | **-11.95** | -0.27 | 12.20 | -0.48 | 0.78 | **-14.12** | -0.27 | 14.37 | -0.48 | 0.78 |
| | S | -0.15 | **2.12** | -0.41 | -4.15 | 2.85 | -0.15 | **9.90** | -0.41 | -7.48 | -1.60 | -0.15 | **6.56** | -0.41 | -5.26 | -0.49 |
| | A | 4.71 | -0.22 | **-3.71** | -0.94 | 0.41 | 6.89 | -0.22 | **-3.71** | -0.94 | -1.76 | 10.15 | -0.22 | **-9.14** | -0.94 | 0.41 |
| | N | 0.90 | -0.40 | -1.73 | **-2.89** | 4.34 | -0.19 | 2.86 | -1.73 | **-2.89** | 2.17 | -0.19 | 2.86 | -1.73 | **-1.80** | 1.08 |
| | F | 1.78 | -1.88 | -8.83 | -0.41 | **9.82** | 8.30 | -0.79 | -7.74 | 5.03 | **-4.31** | 0.69 | -2.96 | -7.74 | -2.58 | **13.08** |
| MV | H | **0.64** | -0.26 | 1.14 | -1.10 | -0.35 | **1.73** | -0.26 | 0.06 | -1.10 | -0.35 | **0.64** | -0.26 | 1.14 | -1.10 | -0.35 |
| | S | -0.15 | **-1.85** | -0.32 | -0.38 | 2.70 | -0.15 | **-7.84** | -0.32 | -1.47 | 9.78 | -0.15 | **-0.95** | -0.32 | -1.47 | 2.89 |
| | A | 5.11 | -0.18 | **-4.03** | -0.36 | -0.40 | 8.38 | -0.18 | **-8.38** | 0.73 | -0.40 | 5.11 | -0.18 | **-4.03** | -0.36 | -0.40 |
| | N | -0.25 | -0.75 | 0.07 | **0.54** | 0.66 | -0.25 | 1.42 | -1.02 | **-0.55** | 0.66 | -0.25 | 0.34 | -1.02 | **1.62** | -0.43 |
| | F | -0.14 | 2.73 | -0.40 | 0.87 | **-2.98** | -0.14 | 3.88 | -0.40 | -0.22 | **-3.04** | -0.14 | 1.68 | -0.40 | -0.22 | **-0.85** |
| MM | H | **-17.62** | -0.26 | 14.60 | -1.46 | 4.81 | **-10.01** | -0.26 | 6.99 | 1.80 | 1.55 | **-11.09** | -0.26 | 11.34 | -0.38 | 0.47 |
| | S | -0.15 | **-7.28** | -0.22 | 3.82 | 3.94 | -0.15 | **-5.23** | -0.22 | 3.94 | 1.76 | -0.15 | **-7.51** | -0.22 | 1.67 | 6.31 |
| | A | 11.74 | 1.01 | **-13.97** | -0.48 | 1.88 | 10.89 | -0.07 | **-13.19** | 1.74 | 0.82 | 12.00 | -0.07 | **-15.42** | 0.63 | 3.04 |
| | N | 1.09 | 0.82 | -1.57 | **-0.16** | -0.11 | 0.00 | 1.91 | -1.57 | **-0.16** | -0.11 | 1.09 | 0.82 | -1.57 | **-1.24** | 0.98 |
| | F | 10.00 | 3.37 | 3.01 | -1.11 | **-15.16** | 21.95 | 1.19 | 9.53 | 2.15 | **-34.72** | 8.91 | 1.19 | 1.92 | 1.07 | **-12.99** |
| SZ | H | **-13.34** | -0.21 | 8.00 | -0.36 | 6.03 | **-9.00** | -0.21 | 4.74 | -0.36 | 4.94 | **-10.08** | -0.21 | 4.74 | -0.36 | 6.03 |
| | S | -0.04 | **-0.37** | -0.04 | 0.73 | -0.11 | -0.04 | **-0.37** | -0.04 | 0.73 | -0.11 | -0.04 | **-3.63** | -0.04 | 1.81 | 2.07 |
| | A | 7.18 | 1.09 | **-6.91** | -1.14 | -0.07 | 10.44 | 0.00 | **-16.70** | 1.03 | 5.36 | 10.44 | 0.00 | **-15.61** | 1.03 | 4.28 |
| | N | 1.05 | 0.48 | 0.37 | **-1.76** | -0.04 | 1.05 | 2.65 | -2.89 | **-2.85** | 2.14 | -0.04 | 0.48 | -0.72 | **-2.85** | 3.23 |
| | F | 9.10 | -0.60 | 0.05 | 3.11 | **-11.45** | 14.54 | 1.57 | 7.66 | 5.29 | **-28.85** | 3.67 | 1.57 | 2.23 | 0.94 | **-8.19** |
| OK | H | **-0.82** | -0.32 | -2.57 | -0.17 | 4.10 | **-6.25** | -0.32 | 5.04 | -0.17 | 1.92 | **-7.34** | -0.32 | 9.38 | -0.17 | -1.34 |
| | S | -0.07 | **0.42** | 1.02 | -0.85 | -0.48 | -0.07 | **-1.76** | -0.07 | 3.50 | -1.57 | -0.07 | **-3.93** | -0.07 | 5.67 | -1.57 |
| | A | -1.59 | -0.14 | **-0.89** | -0.25 | 3.01 | 13.63 | -0.14 | **-15.02** | 0.84 | 0.84 | 13.63 | -0.14 | **-13.93** | -0.25 | 0.84 |
| | N | -0.21 | -7.18 | 0.34 | **6.39** | 0.70 | -0.21 | 10.21 | -0.74 | **-8.83** | -0.39 | -0.21 | 1.51 | -0.74 | **-0.14** | -0.39 |
| | F | -2.29 | -1.31 | 4.94 | 26.62 | **-27.64** | -1.20 | -0.22 | -0.50 | -0.55 | **2.80** | 0.98 | -1.31 | -0.50 | -0.55 | **1.71** |
| BM | H | **-2.44** | -0.56 | 2.82 | 1.50 | -1.10 | **-4.62** | -0.56 | 5.00 | -0.68 | 1.08 | **-1.36** | -0.56 | 3.91 | -0.68 | -1.10 |
| | S | 0.56 | **-3.40** | -0.64 | 3.24 | 0.46 | -0.53 | **3.12** | -0.64 | -1.10 | -0.63 | 0.56 | **-1.23** | -0.64 | 2.16 | -0.63 |
| | A | 2.18 | -0.56 | **-2.28** | 1.45 | -0.63 | 3.27 | -0.56 | **-7.72** | 3.62 | 1.55 | 2.18 | -0.56 | **-4.46** | 3.62 | -0.63 |
| | N | -0.96 | 1.65 | -0.65 | **0.51** | -0.48 | 0.13 | 2.74 | -1.74 | **-1.66** | 0.60 | 0.13 | 3.83 | -2.82 | **-1.66** | 0.60 |
| | F | 0.41 | 0.37 | -0.52 | 0.59 | **-0.64** | -0.67 | 3.63 | -0.52 | -0.49 | **-1.72** | -0.67 | 2.55 | -0.52 | -0.49 | **-0.64** |

TABLE 5
OVERALL CONTRASTIVE RESULTS FOR ALL SPEAKERS FOR THE GEES CORPUS

| Speakers | Emotions | Machine recognition rate in % | | | | | Machine recognition rate in % | | | | | Machine recognition rate in % | | | | |
|---|---|---|---|---|---|---|---|---|---|---|---|---|---|---|---|---|
| | | Utterance-based representations | | | | | Segment-based representations | | | | | Combination-based representations | | | | |
| | | H | S | A | N | F | H | S | A | N | F | H | S | A | N | F |
| All | H | **-7.05** | -0.31 | 5.49 | -0.35 | 2.38 | **-6.69** | -0.31 | 5.67 | -0.17 | 1.65 | **-7.23** | -0.31 | 7.48 | -0.53 | 0.75 |
| | S | 0.00 | **-1.73** | -0.10 | 0.40 | 1.56 | -0.18 | **-0.37** | -0.28 | -0.31 | 1.27 | 0.00 | **-1.78** | -0.28 | 0.76 | 1.43 |
| | A | 4.89 | 0.17 | **-5.30** | -0.29 | 0.70 | 8.91 | -0.20 | **-10.79** | 1.17 | 1.07 | 8.92 | -0.20 | **-10.43** | 0.62 | 1.26 |
| | N | 0.27 | -0.90 | -0.53 | **0.44** | 0.84 | 0.09 | 3.63 | -1.61 | **-2.82** | 0.84 | 0.09 | 1.64 | -1.43 | **-1.01** | 0.84 |
| | F | 3.14 | 0.45 | -0.29 | 4.95 | **-8.00** | 7.13 | 1.54 | 1.34 | 1.87 | **-11.64** | 2.24 | 0.45 | -0.83 | -0.31 | **-1.31** |

Some observations are made from contrastive results in Tables 4 and 5. For all three representations, "happy" is not confused at all with "sad" by our systerm while listeners misclassify some utternaces of "happy" state to be "sad". In addition, the performance of listeners is always superior to their machine counterpart in recognizing some emotional states, e.g., "anger", regardless of speakers. Finally, it is evident that the misclassification by our system often lies in between "happy" and "anger" as well as between "sad" and "neutral". To a lesser extent, the same conclusion can be drawn for "happy" and "fear". But there is little confusion between "sad" and "fear". In addition, "fear" is also misclassified as "neutral". These results are consistent for all three acoustic representations, which highlights the reliability of our system. Apparently, most of the results by our system are consistent with the human listening test in [3].

Fig. 6 summarizes comparative results. In general, our system achieves an overall emotion recognition rate of 90.63% for all speakers by the utterance-based representation. The recognition rate with the segment-based representation comes out to be 88.49%. For combination-based representation, the emotions are recognized with a rate of 90.61% as compared to 95.00% by human listening. Here, all three representations yield almost the same performance, which suggests that emotions are identified well both globally and locally in the utterances of the GEES corpus. The recognition rate for the utterance-based representation on individual corpora ranges from 85.15% to 95.43%. For the segment-based representation, the range is from 83.26% to 93.91%. For combination-based representation, the range is from 86.34% to 96.26%. In contrast, the performance of human listeners ranges from 88.67% to 96.98%. It implies that the sophisticated human auditory system better captures emotional characteristics from speech that conveys mixing types of information.

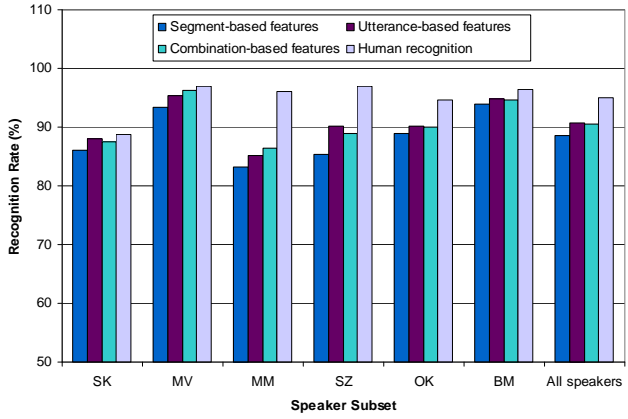

Fig. 6. Contrastive results on individual and whole corpus for the GEES.

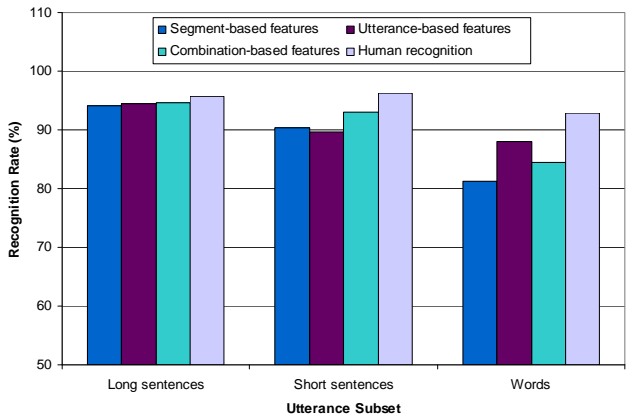

Fig. 7. Contrastive results on utterance sub-corpora for the GEES.

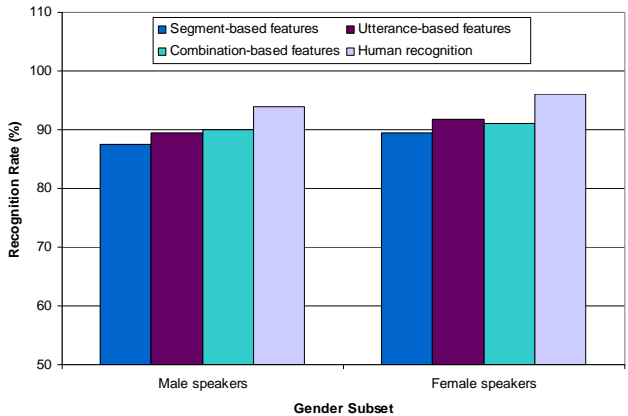

Fig. 8. Contrastive results on gender sub-corpora for the GEES.

We further investigate the performance in terms of length of the utterances. From Fig. 7, it is evident that our system performs slightly better for long sentences than for short sentences. It might suggest that our automatic categorization system does not work well unless sufficient information can be captured from an utterance. In contrast, human listening experiments suggest that the emotional states are recognized relatively regardless of the length of sentences as human is generally good at exploiting various information sources on a fine scale. Since a single word often conveys the limited information by using acoustic features only, it is not surprising to see that the performance for a single-word utterance is inferior to those of sentences for all three feature extraction approaches. It appears sensible that machine performance for words and short sentences is inferior to human listening results.

Finally, we report the performance on gender sub-corpora in Fig. 8 that depicts the recognition rate by collecting the averaging gender results achieved by our system and human listening test. From Fig. 8, we observe that the recognition rate on the female speaker sub-corpus is slightly higher than that on the male speaker sub-corpus for all three representations. This finding has been confirmed by human listeners as shown in Fig. 8. The consistent results on the gender sub-corpora suggest that in the GEES, for both machine and humans, the emotional states in speech are better recognized from the utterances of female speakers than those of male speakers. Although the difference is not statistically significant, the results are interesting for further examination.

### 5.1.2 DES Corpus

The human listening test results for the DES corpus reveal that "happy" and "surprise" states are often confused with each other and so are "sad" and "neutral" states.

The machine recognition rate in the form of confusion matrices for all four speakers on the overall corpus are presented in Tables 6 and 7, respectively. Unfortunately, the detailed human listening results for individual speakers were not reported in [4]. Thus, it is impossible to compare on an individual basis. Given that the recognition rate for every speaker sub-corpora and the confusion matrix of whole corpus were reported in [4], comparative results on the same settings are shown in Table 8.

In Table 8, for our system, all the emotional states are confused with each other to some extent for both utterance-based and segment-based representations. This observation can also be seen in human recognition results [4]. By using the combination-based representation, however, "surprise" is no longer confused at all with "sad" and "neutral", and vice versa. Emotional state pairs mostly misclassified are "happy" and "surprise" as well as "sad" and "neutral" regardless of representations. In addition, there is also confusion between "anger" and "happy" to a lesser extent. As a result, the machine recognition results are highly consistent with the human listening test results

TABLE 6
EMOTION RECOGNITION RATES FOR INDIVIDUAL SPEAKERS FOR THE DES CORPUS

| Speakers | Emotions | Machine recognition rate in % | | | | | Machine recognition rate in % | | | | | Machine recognition rate in % | | | | |
|---|---|---|---|---|---|---|---|---|---|---|---|---|---|---|---|---|
| | | Utterance-based representations | | | | | Segment-based representations | | | | | Combination-based representations | | | | |
| | | H | S | A | N | Sr | H | S | A | N | Sr | H | S | A | N | Sr |
| HO | H | **7.69** | 7.69 | 0.00 | 23.08 | 61.54 | **38.46** | 0.00 | 0.00 | 7.69 | 53.85 | **15.39** | 0.00 | 30.77 | 7.69 | 46.15 |
| | S | 0.00 | **100.00** | 0.00 | 0.00 | 0.00 | 7.69 | **76.92** | 0.00 | 7.69 | 7.69 | 7.69 | **76.92** | 7.69 | 7.69 | 0.00 |
| | A | 50.00 | 8.33 | **8.33** | 16.67 | 16.67 | 41.67 | 8.33 | **0.00** | 8.33 | 41.67 | 8.33 | 16.67 | **25.00** | 8.33 | 41.67 |
| | N | 23.08 | 0.00 | 0.00 | **76.92** | 0.00 | 38.46 | 0.00 | 0.00 | **61.54** | 0.00 | 0.00 | 0.00 | 15.39 | **84.62** | 0.00 |
| | Sr | 53.85 | 0.00 | 7.69 | 0.00 | **38.46** | 30.77 | 0.00 | 0.00 | 0.00 | **69.23** | 38.46 | 0.00 | 0.00 | 0.00 | **61.54** |
| JZB | H | **38.46** | 7.69 | 0.00 | 7.69 | 46.15 | **61.54** | 7.69 | 0.00 | 0.00 | 30.77 | **61.54** | 0.00 | 0.00 | 7.69 | 30.77 |
| | S | 0.00 | **100.00** | 0.00 | 0.00 | 0.00 | 0.00 | **76.92** | 0.00 | 23.08 | 0.00 | 0.00 | **100.00** | 0.00 | 0.00 | 0.00 |
| | A | 0.00 | 0.00 | **100.00** | 0.00 | 0.00 | 50.00 | 0.00 | **50.00** | 0.00 | 0.00 | 8.33 | 0.00 | **91.67** | 0.00 | 0.00 |
| | N | 0.00 | 0.00 | 0.00 | **92.31** | 7.69 | 0.00 | 7.69 | 0.00 | **92.31** | 0.00 | 7.69 | 7.69 | 0.00 | **84.62** | 0.00 |
| | Sr | 46.15 | 7.69 | 0.00 | 0.00 | **46.15** | 41.67 | 8.33 | 0.00 | 0.00 | **50.00** | 41.67 | 0.00 | 0.00 | 0.00 | **58.33** |
| DHC | H | **46.15** | 0.00 | 23.08 | 0.00 | 30.77 | **76.92** | 0.00 | 7.69 | 0.00 | 15.39 | **61.54** | 0.00 | 15.39 | 0.00 | 23.08 |
| | S | 0.00 | **46.15** | 0.00 | 53.85 | 0.00 | 7.69 | **23.08** | 0.00 | 69.23 | 0.00 | 0.00 | **69.23** | 0.00 | 30.77 | 0.00 |
| | A | 7.69 | 0.00 | **76.92** | 0.00 | 15.39 | 30.77 | 0.00 | **69.23** | 0.00 | 0.00 | 15.39 | 0.00 | **76.92** | 0.00 | 7.69 |
| | N | 0.00 | 53.85 | 7.69 | **38.46** | 0.00 | 7.69 | 53.85 | 7.69 | **30.77** | 0.00 | 0.00 | 38.46 | 0.00 | **61.54** | 0.00 |
| | Sr | 46.15 | 0.00 | 0.00 | 0.00 | **53.85** | 23.08 | 0.00 | 0.00 | 0.00 | **76.92** | 53.85 | 0.00 | 0.00 | 0.00 | **46.15** |
| KLA | H | **61.54** | 7.69 | 23.08 | 0.00 | 7.69 | **76.92** | 0.00 | 0.00 | 0.00 | 23.08 | **46.15** | 0.00 | 30.77 | 0.00 | 23.08 |
| | S | 7.69 | **38.46** | 0.00 | 53.85 | 0.00 | 0.00 | **53.85** | 0.00 | 38.46 | 7.69 | 0.00 | **53.85** | 0.00 | 46.15 | 0.00 |
| | A | 69.23 | 0.00 | **0.00** | 15.39 | 15.39 | 69.23 | 0.00 | **0.00** | 0.00 | 30.77 | 23.08 | 0.00 | **38.46** | 23.08 | 15.39 |
| | N | 0.00 | 53.85 | 0.00 | **46.15** | 0.00 | 7.69 | 15.39 | 0.00 | **30.77** | 46.15 | 0.00 | 46.15 | 0.00 | **53.85** | 0.00 |
| | Sr | 15.39 | 0.00 | 0.00 | 0.00 | **84.62** | 23.08 | 7.69 | 0.00 | 0.00 | **69.23** | 7.69 | 0.00 | 7.69 | 0.00 | **84.62** |

TABLE 7
EMOTION RECOGNITION RATES FOR ALL SPEAKERS FOR THE DES CORPUS

| Speakers | Emotions | Machine recognition rate in % | | | | | Machine recognition rate in % | | | | | Machine recognition rate in % | | | | |
|---|---|---|---|---|---|---|---|---|---|---|---|---|---|---|---|---|
| | | Utterance-based representations | | | | | Segment-based representations | | | | | Combination-based representations | | | | |
| | | H | S | A | N | Sr | H | S | A | N | Sr | H | S | A | N | Sr |
| All | H | **38.46** | 5.77 | 11.54 | 7.69 | 36.54 | **63.46** | 1.92 | 1.92 | 1.92 | 30.77 | **46.15** | 0.00 | 19.23 | 3.85 | 30.77 |
| | S | 1.92 | **71.15** | 0.00 | 26.92 | 0.00 | 3.85 | **57.69** | 0.00 | 34.62 | 3.85 | 1.92 | **75.00** | 1.92 | 21.15 | 0.00 |
| | A | 31.73 | 2.08 | **46.31** | 8.01 | 11.86 | 47.92 | 2.08 | **29.81** | 2.08 | 18.11 | 13.78 | 4.17 | **58.01** | 7.85 | 16.19 |
| | N | 5.77 | 26.92 | 1.92 | **63.46** | 1.92 | 13.46 | 19.23 | 1.92 | **53.85** | 11.54 | 1.92 | 23.08 | 3.85 | **71.15** | 0.00 |
| | Sr | 40.39 | 1.92 | 1.92 | 0.00 | **55.77** | 29.65 | 4.01 | 0.00 | 0.00 | **66.35** | 35.42 | 0.00 | 1.92 | 0.00 | **62.66** |

TABLE 8
OVERALL CONTRASTIVE RESULTS FOR ALL SPEAKERS FOR THE DES CORPUS

| Speakers | Emotions | Machine recognition rate in % | | | | | Machine recognition rate in % | | | | | Machine recognition rate in % | | | | |
|---|---|---|---|---|---|---|---|---|---|---|---|---|---|---|---|---|
| | | Utterance-based representations | | | | | Segment-based representations | | | | | Combination-based representations | | | | |
| | | H | S | A | N | Sr | H | S | A | N | Sr | H | S | A | N | Sr |
| All | H | **-17.94** | 4.07 | 7.74 | -0.61 | 6.74 | **7.06** | 0.22 | -1.88 | -6.38 | 0.97 | **-10.25** | -1.70 | 15.43 | -4.45 | 0.97 |
| | S | 1.82 | **-14.05** | -0.30 | 14.32 | -1.80 | 3.75 | **-27.51** | -0.30 | 22.02 | 2.05 | 1.82 | **-10.20** | 1.62 | 8.55 | -1.80 |
| | A | 27.23 | 0.38 | **-28.79** | -2.19 | 3.36 | 43.42 | 0.38 | **-45.29** | -8.12 | 9.61 | 9.28 | 2.47 | **-17.09** | -2.35 | 7.69 |
| | N | 5.67 | -4.78 | -2.88 | **2.66** | -0.68 | 13.36 | -12.47 | -2.88 | **-6.95** | 8.94 | 1.82 | -8.62 | -0.95 | **10.35** | -2.60 |
| | Sr | 11.68 | 0.92 | 0.62 | -10.00 | **-3.33** | 0.95 | 3.01 | -1.30 | -10.00 | **7.25** | 6.72 | -1.00 | 0.62 | -10.00 | **3.56** |

in [4].

The performance of our system against that of humans is shown in Fig. 9. For the utterance-based representation, the average recognition rate for the whole corpus is 55.08% while the recognition rate is 54.55% as the segment-based representation is employed. For combination-based representation, the averaging recognition rate is 62.74%. In contrast, human listeners are able to identify emotional states with a rate of 67.31% on average. For individual speaker sub-corpora, the performance of our system ranges from 46.15% to 75.00% for utterance-based representation, from 46.15% to 66.67% for segment-based representation and from 53.12% to 79.37% for combination-based representation, respectively. In contrast, the human performance varies from 63.08% to 72.31%. It is noticed that the recognition rate of our system for one speaker corpus named JZB is significantly higher as compared to that of other speaker corpora. This result is entirely consistent with the human listening test in [4].

The performance is further evaluated in terms of the utterance length for the DES corpus, as shown in Fig. 10. By using the utterance-based and the combination-based representations, our system achieves a high recognition rate for sentences and words, whereas the recognition rate for passages is relatively low. For the segment-based representation, the recognition rate is higher for words as compared to sentences and passages, which conforms to the nature of the segment-based features that characterizes local critical features in a word. In the listening test [4], humans are able to recognize the passages much better than our system, which may attribute to human's capability of capturing the contextual information that our system does not explore. Nevertheless, the performance of our system for words and sentences, to a great extent, are consistent with the human listening test [4].

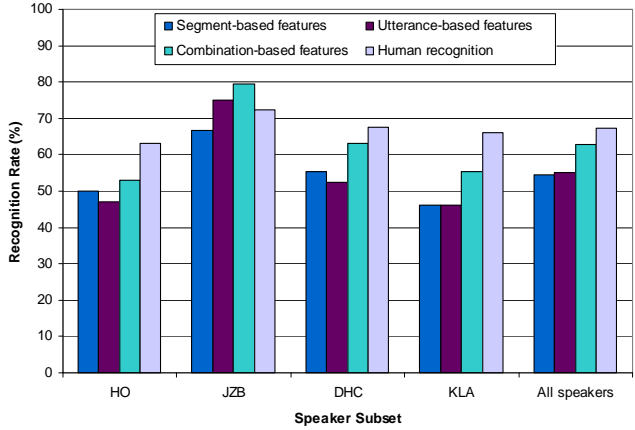

Fig. 9. Contrastive results on individual and whole corpus for the DES.

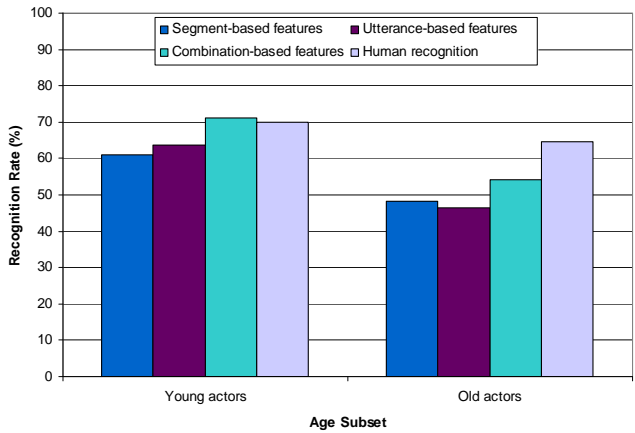

Fig. 10. Contrastive results on utterance sub-corpora for the DES.

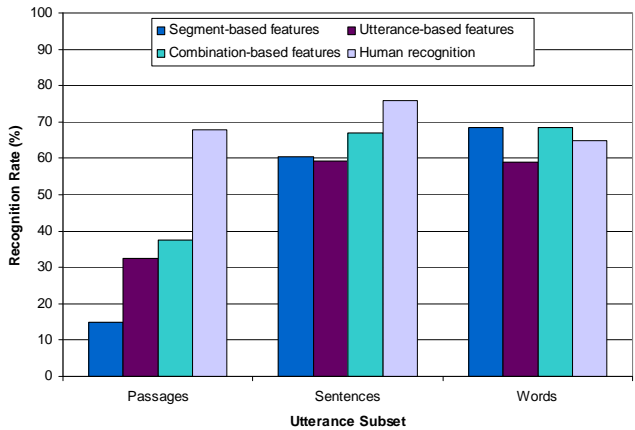

Fig. 11. Contrastive results on actor's age sub-corpora for the DES.

We further investigate the performance in terms of age of the actors in order to make a direct comparison with that reported in [4]. Fig. 11 illustrates the averaging recognition rate over old and young actors achieved by our system and the human listening test. It is evident that the recognition rate of our system is higher for young actors as compared to old ones, which is consistent with the human performance. Results of our system and human listening test suggest that the young actors conveyed the emotions in a better way than the old actors for the DES corpus.

### 5.2 Results on the Universal Feature Subset

Using exactly the same experimental settings on the universal feature subsets listed in Table 1, we achieve results for a scenario regardless of linguistic and semantic features for the two corpora.

#### 5.2.1 GEES Corpus

Recognition rates in the form of confusion matrices for individual and all speaker corpora are reported in Tables 9 and 10, respectively. The results indicate that emotional state pairs often confused with each other are "happy" and "anger" as well as "sad" and "neutral". It is apparent that the results here are consistent with what we have observed as the full feature sets are used and the human listening test in [3].

The comparison of the recognition rate on the full feature set with that on the universal feature subsets is shown in Fig. 12 and 13, respectively. With the utterance-based representation, the recognition rate achieved by using the universal feature subset is almost same as that with the full feature set, as shown in Fig. 12. The recognition rate for the all speaker corpus with the full feature set is 90.63% whereas the recognition rate with the universal feature subset reaches 90.96%. By the use of segment-based representation, the averaging recognition rate on the universal feature subset is nearly the same as that on the full feature set given the fact that the recognition rate for the whole corpus is 88.49% with the full feature set and 87.44% with the universal feature subset. It demonstrates that emotional states can be identified without the use of any semantic cues.

#### 5.2.2 DES Corpus

The recognition results with the universal feature subset in the form of confusion matrices for both individual and whole corpus are shown in Tables 11 and 12, respectively. The confusion often happens between "happy" and "surprise" as well as between "sad" and "neutral". These results are consistent with the results in both machine recognition with the full feature set and human listening test [4].

Fig. 14 and 15 illustrate comparative results on the full feature set and the universal feature subset as the utterance-based and segment-based representations are used, respectively. It is observed from Fig. 14 and 15 that the recognition rate on the universal feature subset is almost as same as that on the full feature set for different speaker corpora regardless of representations. Given the fact that for the whole corpus, the recognition rate is 55.08% for the universal feature subset and 55.81% for the full feature set as the utterance-based representation is used, and

12TABLE 9
EMOTION RECOGNITION RATES FOR INDIVIDUAL SPEAKERS FOR THE GEES CORPUS USING UNIVERSAL FEATURES

| Speakers | Emotions | Machine recognition rate in % | | | | | Machine recognition rate in % | | | | |
|---|---|---|---|---|---|---|---|---|---|---|---|
| | | Utterance-based universal features | | | | | Segment-based universal features | | | | |
| | | H | S | A | N | F | H | S | A | N | F |
| SK | H | **80.44** | 0.00 | 16.30 | 1.09 | 2.17 | **86.96** | 0.00 | 10.87 | 0.00 | 2.17 |
| | S | 0.00 | **96.67** | 1.11 | 1.11 | 1.11 | 0.00 | **72.22** | 0.00 | 25.56 | 2.22 |
| | A | 17.39 | 0.00 | **79.35** | 0.00 | 3.26 | 14.13 | 0.00 | **84.78** | 0.00 | 1.09 |
| | N | 0.00 | 3.26 | 0.00 | **95.65** | 1.09 | 0.00 | 16.30 | 0.00 | **81.52** | 2.17 |
| | F | 2.17 | 1.09 | 1.09 | 3.26 | **92.39** | 11.96 | 3.26 | 2.17 | 10.87 | **71.74** |
| MV | H | **96.74** | 0.00 | 3.26 | 0.00 | 0.00 | **98.91** | 0.00 | 1.09 | 0.00 | 0.00 |
| | S | 0.00 | **92.39** | 0.00 | 1.09 | 6.52 | 0.00 | **75.86** | 0.00 | 20.69 | 3.45 |
| | A | 10.87 | 0.00 | **89.13** | 0.00 | 0.00 | 7.61 | 1.09 | **91.30** | 0.00 | 0.00 |
| | N | 0.00 | 1.09 | 0.00 | **98.91** | 0.00 | 0.00 | 11.96 | 0.00 | **88.04** | 0.00 |
| | F | 0.00 | 10.87 | 0.00 | 1.09 | **88.04** | 0.00 | 6.59 | 0.00 | 2.20 | **91.21** |
| MM | H | **78.26** | 0.00 | 10.87 | 1.09 | 9.78 | **77.17** | 0.00 | 7.61 | 5.44 | 9.78 |
| | S | 1.11 | **90.00** | 0.00 | 7.78 | 1.11 | 0.00 | **51.14** | 0.00 | 45.46 | 3.41 |
| | A | 13.04 | 1.09 | **82.61** | 0.00 | 3.26 | 10.00 | 0.00 | **83.33** | 1.11 | 5.56 |
| | N | 0.00 | 2.17 | 0.00 | **97.83** | 0.00 | 1.09 | 3.26 | 0.00 | **94.57** | 1.09 |
| | F | 5.44 | 3.26 | 5.44 | 1.09 | **84.78** | 16.30 | 1.09 | 10.87 | 6.52 | **65.22** |
| SZ | H | **81.52** | 0.00 | 10.87 | 1.09 | 6.52 | **90.22** | 0.00 | 4.35 | 0.00 | 5.44 |
| | S | 0.00 | **94.57** | 0.00 | 4.35 | 1.09 | 0.00 | **86.96** | 0.00 | 11.96 | 1.09 |
| | A | 14.13 | 0.00 | **81.52** | 0.00 | 4.35 | 10.87 | 1.09 | **73.91** | 1.09 | 13.04 |
| | N | 0.00 | 3.26 | 2.17 | **94.57** | 0.00 | 1.09 | 25.00 | 2.17 | **70.65** | 1.09 |
| | F | 8.70 | 2.17 | 3.26 | 3.26 | **82.61** | 18.48 | 4.35 | 3.26 | 7.61 | **66.30** |
| OK | H | **83.70** | 0.00 | 13.04 | 0.00 | 3.26 | **83.70** | 0.00 | 9.78 | 1.09 | 5.44 |
| | S | 0.00 | **93.48** | 0.00 | 5.44 | 1.09 | 0.00 | **90.22** | 0.00 | 8.70 | 1.09 |
| | A | 13.04 | 0.00 | **84.78** | 0.00 | 2.17 | 22.83 | 0.00 | **77.17** | 0.00 | 0.00 |
| | N | 0.00 | 9.78 | 0.00 | **90.22** | 0.00 | 0.00 | 35.87 | 1.09 | **63.04** | 0.00 |
| | F | 1.09 | 0.00 | 0.00 | 0.00 | **98.91** | 2.17 | 1.09 | 0.00 | 1.09 | **95.65** |
| BM | H | **94.57** | 0.00 | 4.35 | 1.09 | 0.00 | **89.13** | 1.09 | 4.35 | 1.09 | 4.35 |
| | S | 0.00 | **93.48** | 1.09 | 3.26 | 2.17 | 0.00 | **89.13** | 0.00 | 10.87 | 0.00 |
| | A | 1.09 | 0.00 | **94.57** | 2.17 | 2.17 | 5.44 | 0.00 | **88.04** | 2.17 | 4.35 |
| | N | 1.09 | 2.17 | 0.00 | **96.74** | 0.00 | 3.26 | 28.26 | 0.00 | **67.39** | 1.09 |
| | F | 1.09 | 2.17 | 0.00 | 2.17 | **94.57** | 0.00 | 6.52 | 0.00 | 0.00 | **93.48** |

TABLE 10
EMOTION RECOGNITION RATES FOR ALL SPEAKERS FOR THE GEES CORPUS USING UNIVERSAL FEATURES

| Speakers | Emotions | Machine recognition rate in % | | | | | Machine recognition rate in % | | | | |
|---|---|---|---|---|---|---|---|---|---|---|---|
| | | Utterance-based universal features | | | | | Segment-based universal features | | | | |
| | | H | S | A | N | F | H | S | A | N | F |
| All | H | **85.87** | 0.00 | 9.78 | 0.73 | 3.62 | **87.68** | 0.18 | 6.34 | 1.27 | 4.53 |
| | S | 0.19 | **93.43** | 0.37 | 3.84 | 2.18 | 0.00 | **77.59** | 0.00 | 20.54 | 1.88 |
| | A | 11.59 | 0.18 | **85.33** | 0.36 | 2.54 | 11.81 | 0.36 | **83.09** | 0.73 | 4.01 |
| | N | 0.18 | 3.62 | 0.36 | **95.65** | 0.18 | 0.91 | 20.11 | 0.54 | **77.54** | 0.91 |
| | F | 3.08 | 3.26 | 1.63 | 1.81 | **90.22** | 8.15 | 3.82 | 2.72 | 4.71 | **80.60** |

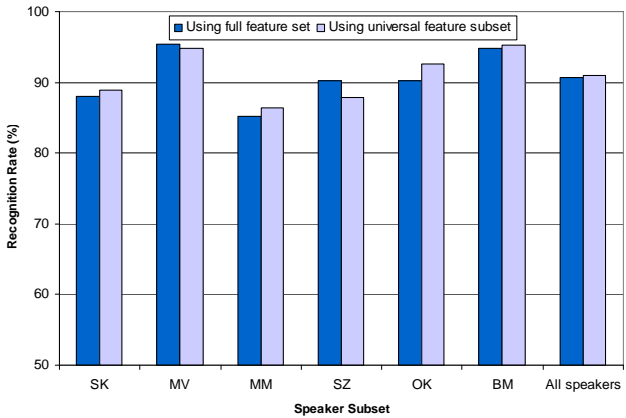

Fig. 12. Utterance-based feature selection contrastive results on individual and whole corpus for the GEES.

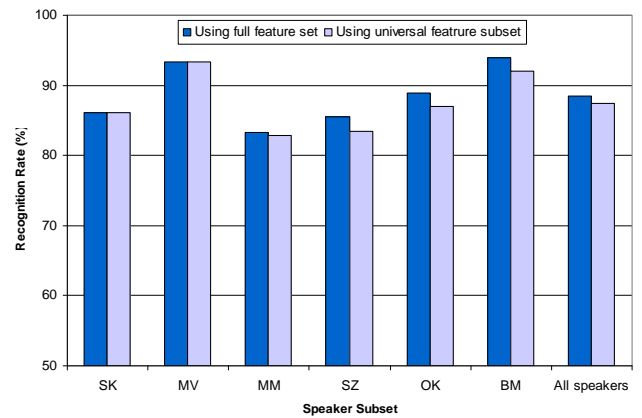

Fig. 13. Segment-based feature selection contrastive results on individual and whole corpus for the GEES.

TABLE 11
EMOTION RECOGNITION RATES FOR INDIVIDUAL SPEAKERS FOR THE DES CORPUS USING UNIVERSAL FEATURES

| Speakers | Emotions | Machine recognition rate in % | | | | | Machine recognition rate in % | | | | |
|---|---|---|---|---|---|---|---|---|---|---|---|
| | | Utterance-based universal features | | | | | Segment-based universal features | | | | |
| | | H | S | A | N | Sr | H | S | A | N | Sr |
| HO | H | **7.69** | 7.69 | 7.69 | 23.08 | 53.85 | **38.46** | 0.00 | 0.00 | 7.69 | 53.85 |
| | S | 0.00 | **76.92** | 0.00 | 15.39 | 7.69 | 0.00 | **92.31** | 0.00 | 7.69 | 0.00 |
| | A | 58.33 | 16.67 | **0.00** | 8.33 | 16.67 | 25.00 | 0.00 | **0.00** | 8.33 | 66.67 |
| | N | 23.08 | 7.69 | 0.00 | **69.23** | 0.00 | 30.77 | 0.00 | 0.00 | **61.54** | 7.69 |
| | Sr | 69.23 | 0.00 | 0.00 | 0.00 | **30.77** | 46.15 | 0.00 | 0.00 | 0.00 | **53.85** |
| JZB | H | **46.15** | 7.69 | 0.00 | 7.69 | 38.46 | **61.54** | 7.69 | 0.00 | 0.00 | 30.77 |
| | S | 0.00 | **84.62** | 0.00 | 7.69 | 7.69 | 15.39 | **46.15** | 0.00 | 38.46 | 0.00 |
| | A | 0.00 | 0.00 | **91.67** | 0.00 | 8.33 | 33.33 | 0.00 | **50.00** | 0.00 | 16.67 |
| | N | 0.00 | 0.00 | 0.00 | **100.00** | 0.00 | 0.00 | 23.08 | 0.00 | **76.92** | 0.00 |
| | Sr | 38.46 | 15.39 | 0.00 | 0.00 | **46.15** | 33.33 | 0.00 | 0.00 | 8.33 | **58.33** |
| DHC | H | **53.85** | 0.00 | 15.39 | 0.00 | 30.77 | **76.92** | 0.00 | 7.69 | 0.00 | 15.39 |
| | S | 0.00 | **61.54** | 0.00 | 38.46 | 0.00 | 7.69 | **53.85** | 0.00 | 38.46 | 0.00 |
| | A | 0.00 | 0.00 | **92.31** | 0.00 | 7.69 | 30.77 | 0.00 | **69.23** | 0.00 | 0.00 |
| | N | 0.00 | 46.15 | 0.00 | **53.85** | 0.00 | 23.08 | 76.92 | 0.00 | **0.00** | 0.00 |
| | Sr | 30.77 | 7.69 | 0.00 | 0.00 | **61.54** | 15.39 | 0.00 | 7.69 | 0.00 | **76.92** |
| KLA | H | **69.23** | 0.00 | 7.69 | 0.00 | 23.08 | **61.54** | 0.00 | 7.69 | 0.00 | 30.77 |
| | S | 7.69 | **38.46** | 0.00 | 53.85 | 0.00 | 7.69 | **69.23** | 0.00 | 23.08 | 0.00 |
| | A | 69.23 | 0.00 | **0.00** | 15.39 | 15.39 | 46.15 | 0.00 | **7.69** | 0.00 | 46.15 |
| | N | 0.00 | 53.85 | 0.00 | **46.15** | 0.00 | 7.69 | 7.69 | 0.00 | **38.46** | 46.15 |
| | Sr | 15.39 | 0.00 | 0.00 | 0.00 | **84.62** | 23.08 | 7.69 | 0.00 | 0.00 | **69.23** |

TABLE 12
EMOTION RECOGNITION RATES FOR ALL SPEAKERS FOR THE DES CORPUS USING UNIVERSAL FEATURES

| Speakers | Emotions | Machine recognition rate in % | | | | | Machine recognition rate in % | | | | |
|---|---|---|---|---|---|---|---|---|---|---|---|
| | | Utterance-based universal features | | | | | Segment-based universal features | | | | |
| | | H | S | A | N | Sr | H | S | A | N | Sr |
| All | H | **44.23** | 3.85 | 7.69 | 7.69 | 36.54 | **59.62** | 1.92 | 3.85 | 1.92 | 32.69 |
| | S | 1.92 | **65.39** | 0.00 | 28.85 | 3.85 | 7.69 | **65.39** | 0.00 | 26.92 | 0.00 |
| | A | 31.89 | 4.17 | **45.99** | 5.93 | 12.02 | 33.81 | 0.00 | **31.73** | 2.08 | 32.37 |
| | N | 5.77 | 26.92 | 0.00 | **67.31** | 0.00 | 15.39 | 26.92 | 0.00 | **44.23** | 13.46 |
| | Sr | 38.46 | 5.77 | 0.00 | 0.00 | **55.77** | 29.49 | 1.92 | 1.92 | 2.08 | **64.58** |

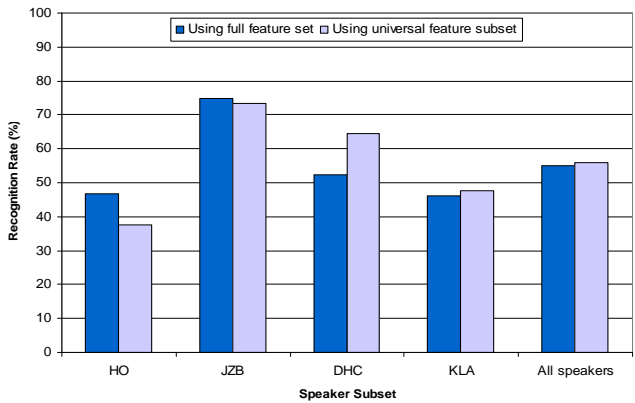

Fig. 14. Utterance-based feature selection contrastive results on individual and whole corpus for the DES.

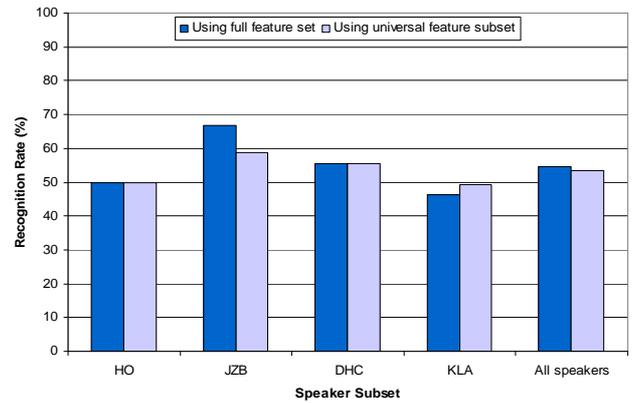

Fig. 15. Segment-based feature selection contrastive results on individual and whole corpus for the DES.

the recognition rate is 53.55% for the universal feature subset and 54.55% for the full feature set as the segment-based representation is used. This outcome is highly consistent with that achieved on the GEES and therefore the same conclusion can be drawn again.

## 6 CONCLUSION

We have presented a systematic investigation for automatic emotional state categorization of speech signals on the GEES and the DES corpora with a psychologically-inspired multistage categorization strategy under the same experimental setup of human listening test. Despite the use of only acoustic features, our system behaves con-sistently with the human listening test and yields performance close to that of humans in most cases. In particular, the performance of using the universal feature subset on the two corpora are competitive with that on the full feature set and that of human listening test. The use of universal feature subset exhibits the same behaviours as the use of the full feature set. As there are no automatic categorization study on the GEES corpus and no comparable study with human listening test on the DES corpus, our work would serve as a baseline in terms of various scenarios for further research in both psychology and automatic emotional state categorization from speech signals.

## ACKNOWLEDGMENT

Authors are grateful to the owners of the GEES and the DES for providing their emotional speech corpora. Authors are also thankful to R. Fernandez for providing his code for feature extraction used in our study. A. Shaukat would like to thank National University of Sciences and Technology (NUST), Pakistan for their financial support.